\documentclass{article}
\pdfoutput=1
\PassOptionsToPackage{numbers, compress}{natbib}
\usepackage[final]{nips_2016}

\usepackage[utf8]{inputenc} 
\usepackage[T1]{fontenc}    
\usepackage{url}            
\usepackage{booktabs}       
\usepackage{amsfonts}       
\usepackage{nicefrac}       
\usepackage{microtype}      
\usepackage{amsmath}
\usepackage{enumerate}
\usepackage{subfigure}
\usepackage{graphicx}
\usepackage{authblk}

\newcommand{\transpose}{^\mathsf{T}}

\title{Deep Learning Approximation for Stochastic \\Control Problems}

\author[1]{Jiequn Han}
\author[1,2,3]{Weinan E}
\affil[1]{The Program of Applied Mathematics, Princeton University}
\affil[2]{School of Mathematical Sciences, Peking University}
\affil[3]{Beijing Institute of Big Data Research}

\begin{document}

\maketitle

\begin{abstract}
Many real world stochastic control problems suffer from the
``curse of dimensionality''. 
To overcome this difficulty, we develop a deep learning approach 
that directly solves high-dimensional stochastic control problems 
based on Monte-Carlo sampling. 
We approximate the time-dependent controls as feedforward neural networks and 
stack these networks together through model dynamics. 
The objective function for the control problem
plays the role of the loss function for the deep neural network. 
We test this approach using examples from the areas of 
optimal trading and energy storage. Our results
suggest that the algorithm presented here achieves satisfactory accuracy and at the 
same time, can handle rather high dimensional problems. 
\end{abstract}

\section{Introduction}
The traditional way of solving stochastic control problems is through
the principle of dynamic programming.  While being mathematically elegant, 
for high-dimensional problems this approach runs into the
technical difficulty 
associated with the ``curse of dimensionality''.  In fact,
it is precisely in this context that 
the term was first introduced, by Richard Bellman \citep{Bellman1957}.
It turns out that the same problem is also at the heart of many other subjects such as machine learning and quantum many-body physics.

In recent years, deep learning has shown impressive results on a variety of
hard problems in machine learning \citep{Lecun1998,Bengio2009,Krizhevsky2012,LeCun2015}, 
suggesting that deep neural networks might
be an effective tool for dealing with the curse of dimensionality problem.
It should be emphasized that although there are partial analytical results,
the reason why deep neural networks have performed so well still largely
remains a mystery.
Nevertheless, it motivates using the deep neural network approximation
in other contexts where curse of dimensionality is the essential obstacle.

In this paper, we develop the deep neural network approximation 
in the context of stochastic control problems.
Even though this is not such an unexpected idea, and has in fact already
been explored in the context of reinforcement learning \citep{Sutton1998}, 
a subject that overlaps substantially with 
control theory, our formulation of the problem still has some merits.  
First of all, the framework we propose here is much simpler than
the corresponding work for reinforcement learning.
Secondly, we study the problem in finite horizon.
This makes the optimal controls time dependent.
Thirdly, instead of formulating approximations to the value function
as is commonly done \citep{Powell2011}, our formulation is in terms
of approximating the optimal control at each time.
In fact, the control at each time step is approximated by a feedforward subnetwork.
We stack these subnetworks together to form a very deep network and train them simultaneously.
Numerical examples in section \ref{sec: numerical} suggest that this approximation
can achieve near-optimality and at the same time handle high-dimensional
problems with relative ease.

We note in passing that 
research on similar stochastic control problems has evolved under the name 
of deep reinforcement learning in the artificial intelligence (AI) community
 \citep{Mnih2015,Silver2016,Schulman2015,Lillicrap2016,JohnSchulman2016}.
As was stressed in \citep{Duan2016}, most of these papers
deal with the infinite horizon problem with time-independent policy.
In contrast, our algorithm only involves a single deep network 
obtained by stacking together, through model dynamics, 
the different subnetwork approximating the time-dependent controls.

In dealing with high-dimensional stochastic control problems, 
the conventional approach taken by the operations research (OR) community has been 
``approximate dynamic programming'' (ADP) \citep{Powell2011}. 
There are two essential steps in ADP. The first is replacing the true value function using some function approximation. 
The second is advancing forward in time from a sample path with backward sweep to update the value function. Unlike ADP, we do not deal with value function at all.
We deal directly with the controls. In addition, our approximation scheme
appears to be more generally applicable.



\section{Mathematical formulation}
We consider a stochastic control problem with finite time horizon $T$
on a probability space $(\Omega,\mathcal{F},P)$ 
with a filtration $\mathcal{F}_0\subset\mathcal{F}_1\subset\cdots\subset\mathcal{F}_T=\mathcal{F}$. Throughout the paper we adopt the convention that any variable indexed by $t$ is $\mathcal{F}_t$-measurable.
We use $s_t\in\mathcal{S}_t\subset\mathbb{R}^m$ to denote the state variable, where $\mathcal{S}_t$ is the set of potential states. The control variable is denoted by $a_t\in\mathbb{R}^n$. 

Our setting is model-based.  We assume that
the evolution of the system is described by
the stochastic model:
\begin{equation}
  s_{t+1} = s_t+b_t(s_t,a_t)+\xi_{t+1}.
  \label{eqn:dynamics}
\end{equation}
Here $b_t$ is the deterministic drift term given by the model. 
$\xi_{t+1}\in\mathbb{R}^{n}$ is a $\mathcal{F}_{t+1}$-measurable random variable
that contains all the noisy information arriving between time $t$ and $t+1$. 
One can view this as a discretized version of stochastic differential equations.
To ensure generality of the model, 
we allow some state-dependent constraints on the control for all $t$:
\begin{align}
  &g_i(s_t,a_t) = 0, \quad\text{for }i=1,\dots,I.\\
  &h_j(s_t,a_t) \ge 0, \quad\text{for }j=1,\dots,J.
\end{align}
Assuming the state variable $s_t$ completely characterizes the model
(in the sense that the optimal control $a_t$ depends only on the current state
$s_t$), 
we can write the set of admissible functions for $a_t$ as
\begin{equation}
  a_t\in\mathcal{A}_t=\left\{a_t(s_t):\mathbb{R}^m\rightarrow \mathbb{R}^n\left|
  \begin{matrix}
    &g_i(s_t,a_t) = 0\text{ for }i=1,\dots,I,~\forall s_t\in\mathcal{S}_t\\
    &h_j(s_t,a_t) \ge 0\text{ for }j=1,\dots,J,~\forall s_t\in\mathcal{S}_t
  \end{matrix}
  \right.\right\}.
\end{equation}

Our problem is finally formulated as (taking minimization for example)
\begin{equation}
  \min_{a_t\in\mathcal{A}_t,~t=0,\cdots,T-1}\mathbb{E}\big\{C_T \mid s_0\big\} = \min_{a_t\in\mathcal{A}_t,~t=0,\cdots,T-1}\mathbb{E}\big\{\sum_{t=0}^{T-1} c_t(s_t,a_t(s_t))+c_T(s_T)\mid s_0\big\},
\end{equation}
where $c_t(s_t,a_t)$ is the intermediate cost, $c_T(s_T)$ is the
final cost and $C_T$ is the total cost.
For later purpose we also define the cumulative cost
\begin{equation}
  C_t = \sum_{\tau=0}^{t}c_\tau(s_\tau,a_\tau),\quad t<T.
\end{equation}


\section{An neural network approximation algorithm}\label{sec:algorithm}

\begin{figure}[ht]
  \centering
  \includegraphics[scale=.72]{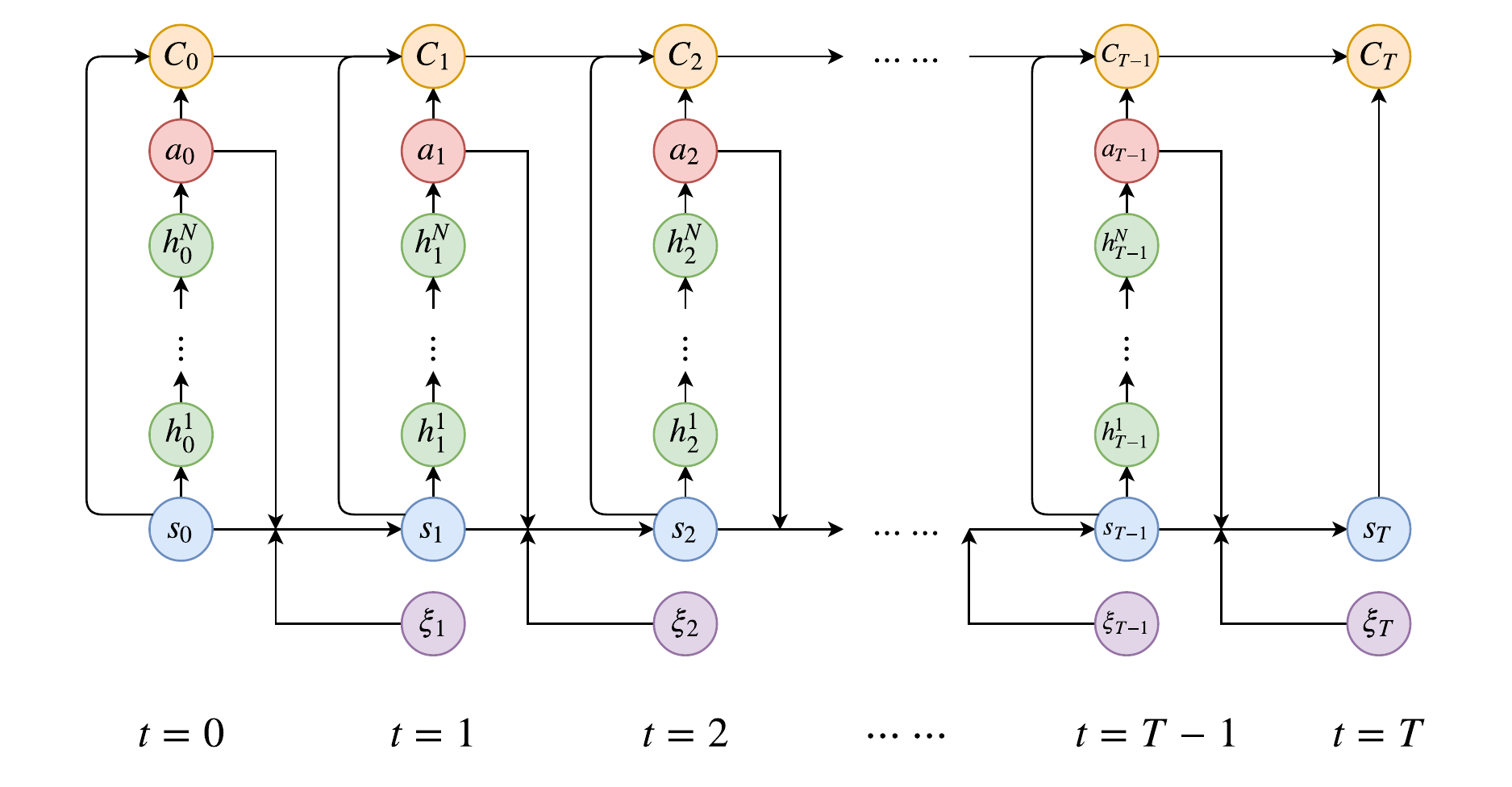}\\
  \caption{Illustration of the
 network architecture for stochastic control problems with $N$ hidden layers for 
each subnetwork. Each column (except $\xi_t$) corresponds to a subnetwork at $t$. 
$h^1_t, ... h^N_t$ are the hidden variables in the subnetwork at time $t$.}
  \label{fig:nn}
\end{figure}

Our task is to approximate the functional dependence of the control on the state,
i.e. $a_t$ as a function of $s_t$. Here we assumed that there are no memory effects
but if necessary, memory effects can also be taken into account with no difference
in principle.
We represent this dependence by a 
multilayer feedforward neural network, 
\begin{equation}
  a_t(s_t)\approx a_t(s_t|\theta_t),
\end{equation}
where $\theta_t$ denotes parameters of the neural network.
Note that we only apply the nonlinear activation function at the layers
for the hidden variables and no activation function is used
at the final layer.
To better explain the algorithm, we assume temporarily 
that there are no other constraints but the
$\mathcal{F}_t$-measurability for the control. 
Then the optimization problem becomes
\begin{equation}
  \min_{\{\theta_t\}_{t=0}^{T-1}}\mathbb{E}\big\{\sum_{t=0}^{T-1}c_t(s_t,a_t(s_t|\theta_t))+c_T(s_T)\}.
\end{equation}
Here for clarity, we ignore the conditional dependence on the initial distribution.
A key observation for the derivation of algorithm is that given a sample of the
stochastic process $\{\xi_t\}_{t=1}^{T}$, 
the total cost $C_T$ can be regarded as the
output of a deep neural network. 
The general architecture of the network is illustrated in Figure \ref{fig:nn}. Note that there are three types of connections in this network:
\begin{enumerate}[\hspace{\parindent}1)]
  \item $s_t\rightarrow h_t^1 \rightarrow h_t^2 \rightarrow \cdots \rightarrow h_t^N \rightarrow a_t$ is the multilayer feedforward neural network approximating 
the control at time $t$. 
The weights $\theta_t$ of this subnetwork are the parameters we aim to optimize.
  \item $(s_t,a_t,C_t)\rightarrow C_{t+1}$ is the direct contribution to the final output of the network. 
Their functional form is determined by the cost function $c_t(s_t,a_t)$. 
There are no parameters to be optimized in this type of connection.
  \item $(s_t,a_t,\xi_{t+1})\rightarrow s_{t+1}$ is the shortcut connecting blocks at different time, which is completely characterized by \eqref{eqn:dynamics}. There are also no parameters to be optimized in this type of connection.
\end{enumerate}
If we use $N$ hidden layers in each subnetwork, 
as illustrated in Figure \ref{fig:nn}, then the whole network has 
$(N+2)T$ layers in total. 

To deal with constraints, we revise the cumulative cost $C_t$ in Figure \ref{fig:nn} by introducing:
\begin{equation}
  C_t \leftarrow C_t+\sum_{i=1}^I \lambda_iP_e(g_i(s_t,a_t)) + \sum_{j=1}^J \sigma_jP_{ie}(h_j(s_t,a_t)),\quad t<T.
\end{equation}
Here $P_e(\cdot),P_{ie}(\cdot)$ are the penalty functions for equality and inequality constraints while $\lambda_i,\sigma_j$ are 
penalty coefficients. Specific examples can be found below.
We should stress that in the testing stage, 
we project the optimal controls we learned to the admissible set 
to ensure that they strictly satisfy all the constraints.

\subsection{Training algorithm}
During training we sample $\{\xi_t\}_{t=1}^{T}$ as the input data and compute $C_T$ from the neural network.
The standard stochastic gradient descent (SGD) method with backpropagation
can be easily adapted to this situation. 
The training algorithm can be easily implemented using common libraries (\textit{e.g.}, TensorFlow \citep{Abadi2015}) without modifying the SGD-type optimizers.
We also adopted the technique of batch normalization \citep{Ioffe2015} 
in the subnetworks, right after each linear transformation and before activation. 
This method accelerates the training by allowing a larger step size and 
easier parameter initialization.

\subsection{Implementation}
We briefly mention some details of the implementation. All our numerical examples are run on a Dell desktop with 3.2GHz Intel Core i7, without any GPU accerleration. We use TensorFlow to implement our algorithm with the Adam optimizer \citep{Kingma2015} to optimize parameters. Adam is an variant of the SGD algorithm, 
based on adaptive estimates of lower-order moments. We set the default values for corresponding hyper-parameters as recommended in \citep{Kingma2015}. 
To deal with the constraints, we choose the quadratic function as penalty functions:
\begin{equation}
  P_e(x) = x^2\quad\text{and}\quad P_{ie}(x) = \min{\{0,x^2\}}.
\end{equation}

For the architecture of the subnetworks, we set the number of layers to 4, 
with 1 input layer (for the state variable), 
2 hidden layers and 1 output layer (representing
the control). 
We choose rectified linear unit (ReLU) as our activation function for the
hidden variables. All the weights in the network are initialized using
a normal distribution without any pre-training.

In the numerical results reported below, 
to facilitate the comparison with the benchmark, we fix the initial state to 
some deterministic value rather than from a random distribution. 
Therefore the optimal control at $t=0$ is also deterministic and batch
normalization is skipped at $t=1$.

\section{Numerical results and discussion}\label{sec: numerical}

\subsection{Execution costs for portfolios}
Our first example is in the area of finance.
It is concerned with minimizing the expected cost for trading blocks of stocks over a fixed time horizon. When a portfolio requires frequent rebalancing, 
large orders across many stocks may appear, which must be executed within a relatively short time horizon. The execution costs associated with such tradings are often substantial, and this calls for smart trading strategies. 
Here we consider a linear percentage price-impact model based on the work of \citep{Bertsimas1998,Bertimas1999}. The reason we choose this example is that it has an analytic solution, which facilitates the evaluation of our numerical solutions.

Denote by $a_t=(a_{1t},a_{2t},\cdots,a_{nt})\transpose\in\mathbb{R}^{n}$ the number of shares of each stock bought in period $t$ at price $p_t=(p_{1t},p_{2t},\cdots,p_{nt})\transpose\in\mathbb{R}^{n}$, $t=0,1,\cdots,T-1$. The investor's objective is  to
\begin{equation}
  \min_{\{a_t\}_{t=0}^{T-1}}\mathbb{E}\sum_{t=0}^{T-1}p_t\transpose a_t,
\end{equation}
subject to $\sum_{t=0}^{T-1}a_t= \bar{a}\in\mathbb{R}^{n}$, where $\bar{a}$ denotes the shares of the
$n$ stocks to be purchased within time $T$. The execution price $p_t$ is assumed to be the sum of two components
\begin{equation*}
  p_t=\tilde{p_t}+\delta_t.
\end{equation*}
Here $\tilde{p}_t$ is the ``no-impact'' price, modeled by geometric Brownian motion, and $\delta_t$ is the impact price, modeled by
\begin{equation}
  \delta_t = \tilde{P}_t(A\tilde{P}_ta_t+Bx_t),
\end{equation}
where $\tilde{P}_t=\text{diag}[\tilde{p}_t]$, $x_t\in\mathbb{R}^{m}$ captures the potential influence of market conditions and $A\in\mathbb{R}^{n\times n}, B\in\mathbb{R}^{n\times m}$. To complete the model specification, we set the dynamics of $x_t$ as a simple multivariate autoregressive process:
\begin{equation}
  x_t = Cx_{t-1} + \eta_t,
\end{equation}
where $\eta_t$ is a white noise vector and $C\in\mathbb{R}^{m\times m}$. The state variable of this model can be chosen as $(\tilde{p}_t, x_t, w_t)$, where $w_t$ denotes the remaining shares to be bought at time $t$. 
This problem can be solved analytically using dynamic programming, see
\citep{Bertimas1999} for the analytic expression of the
optimal execution strategy and the corresponding optimal cost.

\begin{figure}[ht]
  \centering
  \subfigure[relative trading cost]{\includegraphics[scale=.52]{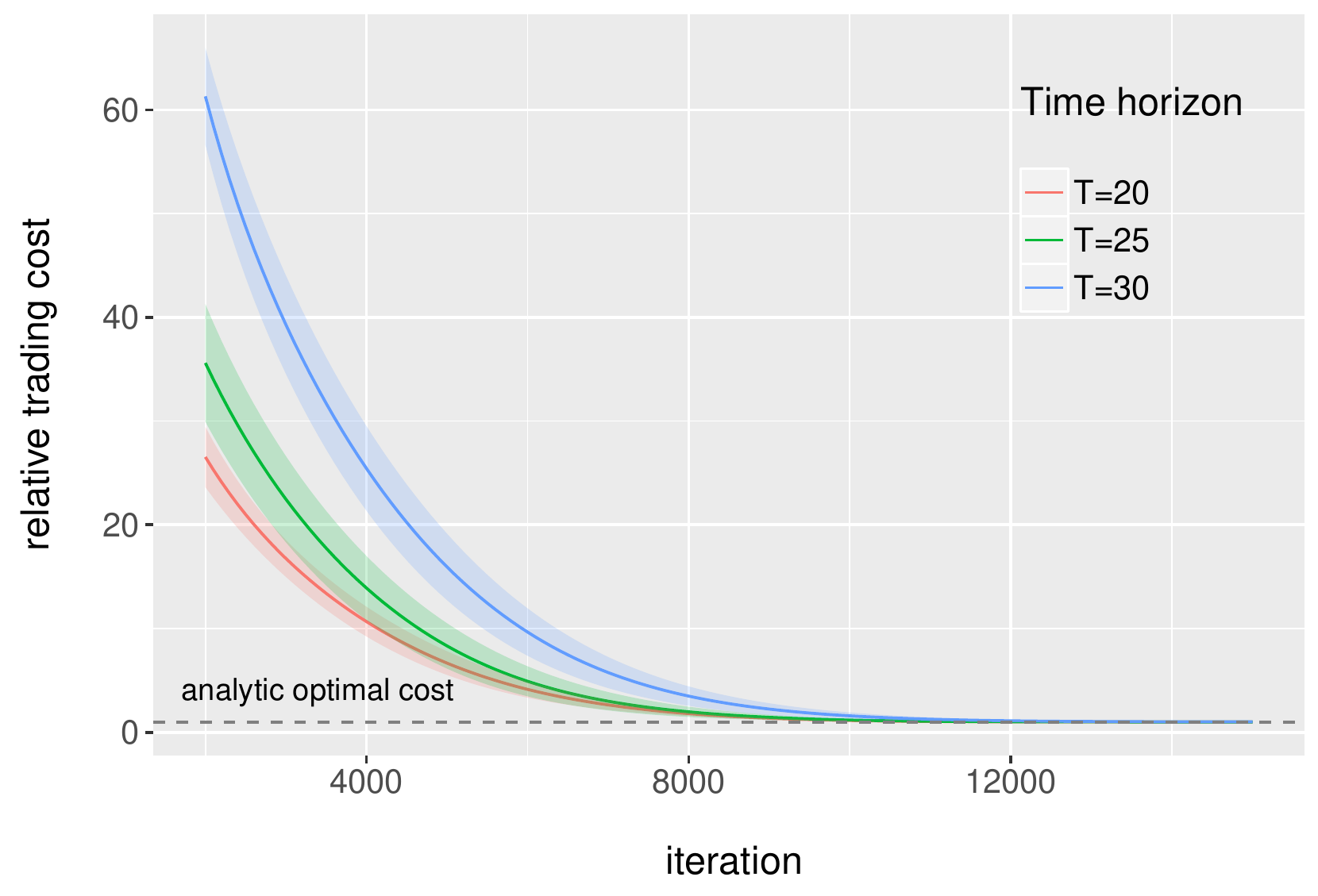}}\\
  \subfigure[relative error for the controls]{\includegraphics[scale=.52]{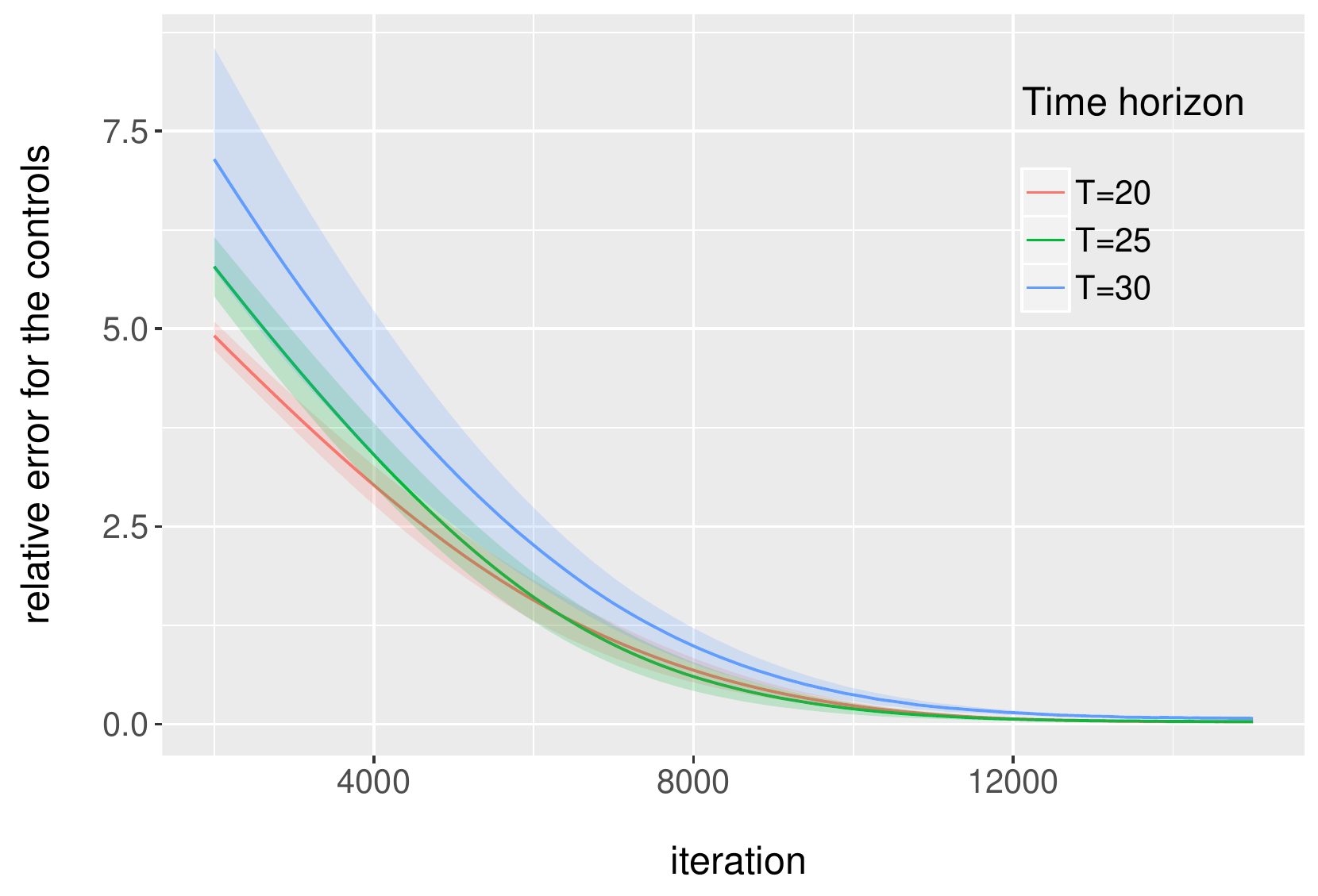}}
  \caption{Relative trading cost and relative error for the controls (compared to the
exact solution) as a function of the number of iterations on validation samples. The shaded area depicts the mean $\pm$ the standard deviation over five different random seeds. The average relative trading cost and relative error for the controls
on test samples are $1.001, 1.002, 1.009$ and $3.7\%, 3.7\%, 8.6\%$ for $T=20, 25, 30$. The average running time is 605~s, 778~s, 905~s respectively.}
  \label{fig:execution}
\end{figure}

In our implementation, all the parameters of the model are assigned with realistic values. We choose $n=10$ and $m=3$, which gives us a generic high-dimensional problem with the control space: $\mathbb{R}^{23}\rightarrow \mathbb{R}^{10}$. We set the number of hidden units in the two hidden layers to 100, the initial learning rate to 0.001, the batch size to 64 and iteration steps to 15000. The learning curves over five different random seeds with different time horizons are plotted in Figure \ref{fig:execution}.

The dashed line in Figure \ref{fig:execution} (a) represents the analytical optimal trading cost (rescaled to 1), defined as the optimal execution cost in cents/share above the no-impact cost $p_0\transpose\bar{a}$. For this problem, the objective function achieves near-optimality with good accuracy:  average relative trading cost to
the exact solution are $1.001, 1.002, 1.009$ for $T=20, 25, 30$. From Figure \ref{fig:execution} (b) we also observe that computed optimal strategy approximates the exact solution well. 
Note that for $T=30$, there are 120 layers in total.

In most practical applications, there are usually constraints on execution strategies. For example, a feasible buying strategy might require $a_t$ to be nonnegative. Such constraints can be imposed easily by adding penalty terms in our algorithm. Here we leave the optimization task with constraints to the next example. 

\subsection{Energy storage and allocation benchmark}
Storage of wind energy has recently received significant attention as a way to increase the efficiency of the electrical grid. Practical and adaptive methods for the optimal energy allocation on such power systems are critical for them to operate reliably and economically. Here we consider an allocation problem from \citep{Salas2013,Jiang2015}, which aims at optimizing revenues from a storage device and a renewable wind energy source while satisfying stochastic demand. 

\begin{figure}[ht]
  \centering
  \includegraphics[scale=.5]{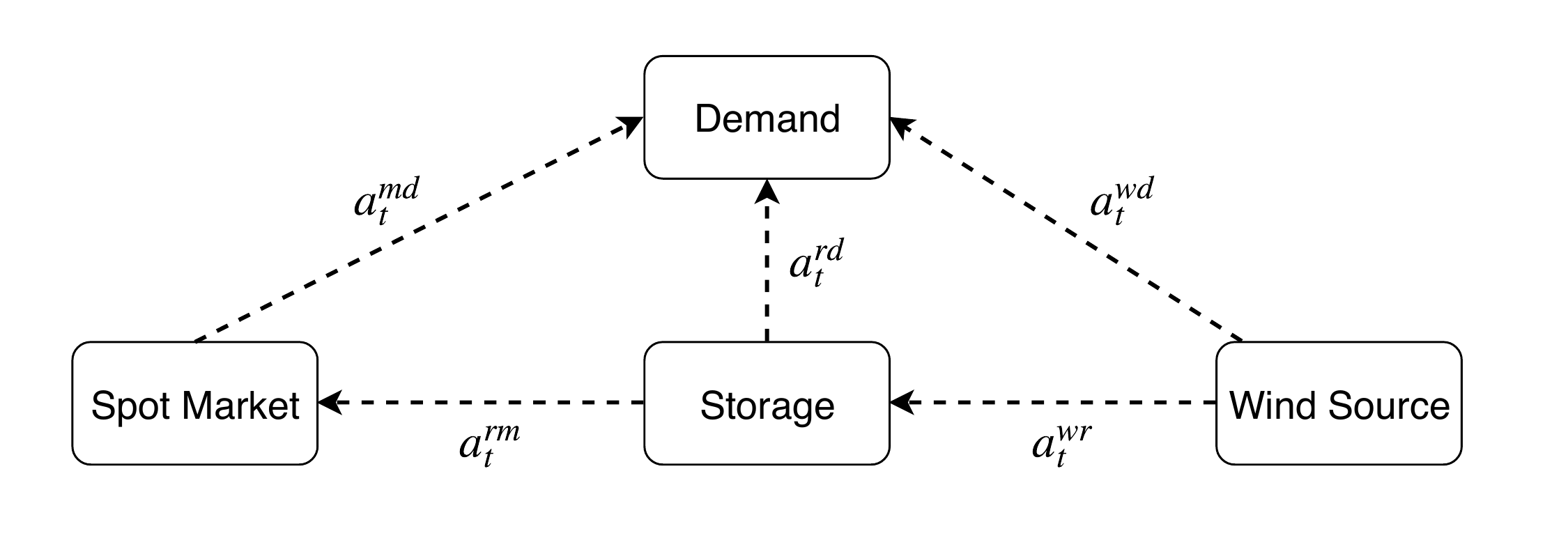}
  \caption{Network diagram for the energy storage problem.}
  \label{fig:energy_network}
\end{figure}

The model is set up as follows. Let the state variable be $s_t=(r_t,w_t,p_t,d_t)$, where $r_t$ is the amount of energy in the storage device, $w_t$ is the amount of energy produced by the wind source, $p_t$ is the price of electricity on the spot market, and $d_t$ is the demand to be satisfied. Let $\gamma^c, \gamma^d$ be the maximum rates of charging and discharging from the storage device respectively, and $r_{\text{max}}$ be the capacity of the storage device. The control variable is given by $a_t=(a_t^{wd}, a_t^{md}, a_t^{rd}, a_t^{wr}, a_t^{rm})$, where $a_t^{ij}$ is the amount of energy transferred from $i$ to $j$ at time $t$. The superscript $w$ stands for wind, $d$ for demand, $r$ for storage and $m$ for spot market. Figure \ref{fig:energy_network} illustrates the meaning of the control components in a network diagram.

We require that all components of $a_t$ be nonnegative. We also require:
\begin{align*}
  a_t^{wd}+a_t^{rd}+a_t^{md} &=d_t,\\
  a_t^{wr}+a_t^{wd} &\leq w_t,\\
  a_t^{rd}+a_t^{rm} &\leq\min\{r_t,\gamma^d\},\\
  a_t^{wr} &\leq\min\{r_{\text{max}}-r_t,\gamma^c\}.
\end{align*}
The intermediate reward function at time $t$ is
\begin{equation}
  c_t(s_t,a_t)=p_t(d_t+a_t^{rm}-a_t^{md}).
\end{equation}
Here we do not consider the holding cost. Let $\phi=(0,0,-1,1,-1)\transpose$. The dynamics for $r_t$ is characterized by
\begin{equation}
  r_{t+1} = r_t+\phi\transpose a_t,
\end{equation}
and $w_t, p_t, d_t$ are modeled by first-order Markov chains in bounded domains, which are all independent from the control (See S2 case in \citep{Jiang2015} for the exact specification).

\begin{figure}[ht]
  \centering
  \includegraphics[scale=.52]{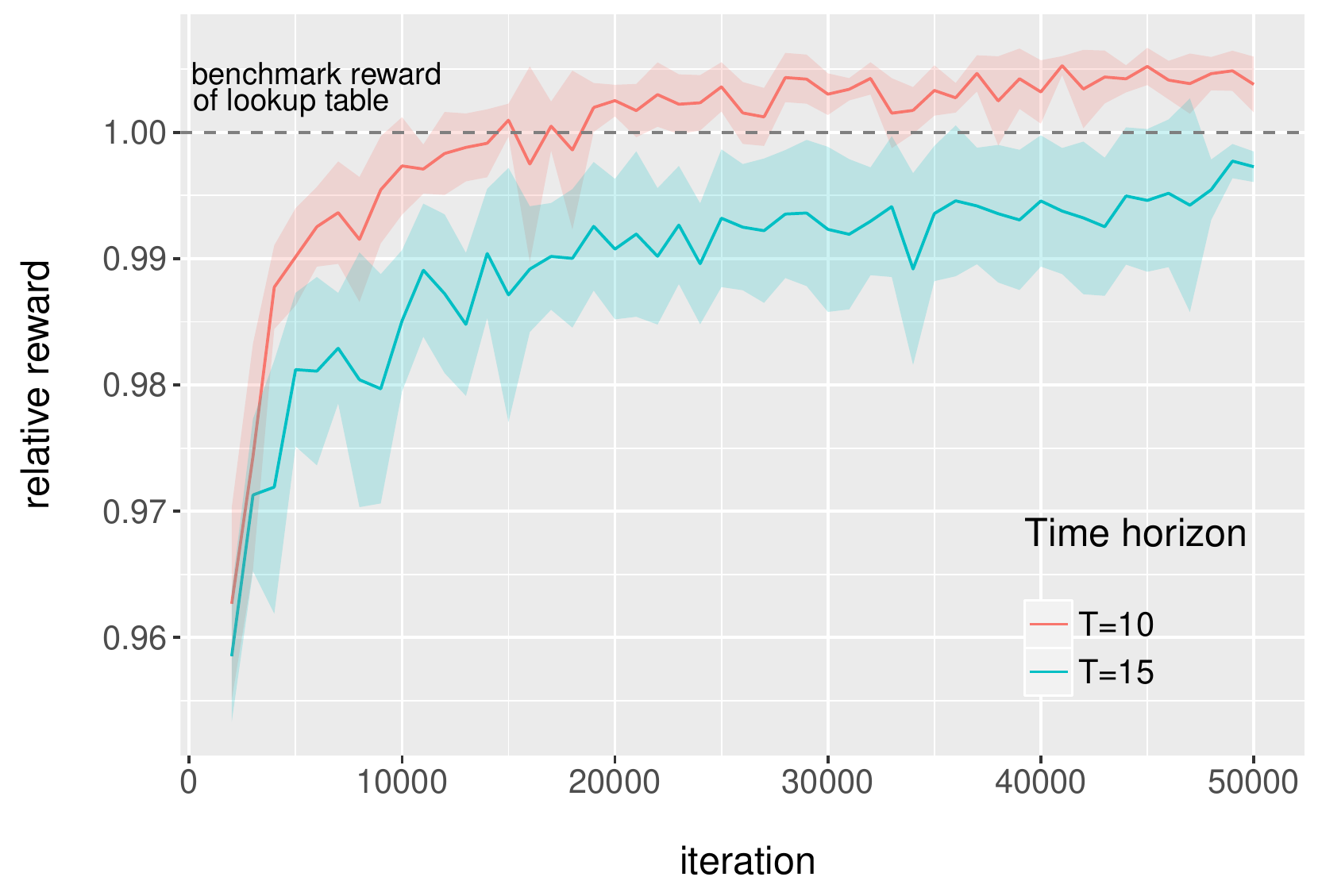}
  \caption{Relative reward as a function of the number of iterations on validation samples, with optimal lookup table obtained from backward dynamic programming being a benchmark. The shaded area depicts the mean $\pm$ the standard deviation over five different random seeds. The average relative reward on test samples is $1.002, 0.995$ for $T=10, 15$. The average running time is 5041~s and 8150~s.}
  \label{fig:single_reward}
\end{figure}

To maximize the total reward, we need to find optimal control in the space $\mathbb{R}^{4}\rightarrow \mathbb{R}^{5}$. Since all the components of control should be negative, we add a ReLU activation at the final layer of each subnetwork. We set the number of hidden units in the two hidden layers to 400, the batch size to 256, all the penalty coefficients to 500 and iteration steps to 50000. The learning rate is 0.003 for the first half of the iterations and 0.0003 for the second half. In the literature, many algorithms for multidimensional stochastic control problems, \textit{e.g.} the ones in \citep{Salas2013,Jiang2015}, proceed by discretizing the state variable and the control variable into a finite set, and present the optimal control in the form of a lookup table. In contrast, our algorithm can handle continuous variables directly. However, for the ease of comparison with the optimal lookup table obtained from backward dynamic programing as in \citep{Jiang2015}, here we artificially confine $w_t, p_t, d_t$ to the values in their lookup table. The relative reward over five different random seeds are plotted in Figure \ref{fig:single_reward}.

 Despite the presence of multiple constraints, our algorithm still gives near-optimal reward. When $T=10$, the neural-network policy gives even higher expected reward than the lookup table policy. It should be noted that if we relax the discretization constraint we imposed on $w_t, p_t, d_t$, then our method can achieve better reward than the lookup table in both cases of $T=10$ and $15$. 

 The learning curves in Figure \ref{fig:single_reward} display clearly a feature of our algorithm for this problem: as time horizon increases, variance becomes larger with the same batch size and more iteration steps are required. We also see that the learning curves are rougher than those in the first example. This might be due to the presence of multiple constraints that result in more nonlinearity in the optimal control policy.

\subsection{Multidimensional energy storage}
Now we extend the previous example to the case of $n$ devices and test the algorithm's performance for the rather high dimensional problems, in which we do not find any other available solution for comparison. We consider the situation of pure arbitrage, \textit{i.e.} $d_t=0$ for all $t$, and allow buying electricity from the spot market to store in the device. The state variable is $s_t=(r_t,w_t,p_t)\in\mathbb{R}^{n+2}$ where $r_t = (r_{1t}, r_{2t},\cdots, r_{nt})\in\mathbb{R}^n$ is the resource vector denoting the storage of each device. The control variable is characterized by $a_t=(a_{1t}^{wr}, a_{1t}^{rm}, a_{1t}^{mr}, \cdots, a_{nt}^{wr}, a_{nt}^{rm}, a_{nt}^{mr})\in\mathbb{R}^{3n}$. $r_{i,\text{max}}, \gamma_i^c, \gamma_i^d$ denote the energy capacity, maximum charging rates and discharging rates of storage device $i$ respectively. We also introduce $\eta_i^c, \eta_i^d$, which are no larger than $1$, as the charging and discharging efficiency of storage device $i$. The holding cost is no longer zero as before, but denoted by $\beta_i$. The intermediate reward function at time $t$ is revised to be
\begin{equation}
  c_t(s_t,a_t)=\sum_{i=1}^n p_t(\eta_i^d a_{it}^{rm}-a_{it}^{mr})-\beta_i{r_{it}},
\end{equation}
and the dynamics for $r_{it}$ becomes
\begin{equation}
  r_{i(t+1)} = r_{it}+\phi_i\transpose a_{it},
\end{equation}
with $\phi_i=(\eta_i^c, -1, \eta_i^c)\transpose$.

We make a simple but realistic assumption that a device with higher energy capacity $r_{i,\text{max}}$ has lower power transfer capacity $\gamma_i^c, \gamma_i^d$, lower efficiency $\eta_i^c, \eta_i^d$ and lower holding cost $\beta_i$. All these model parameters are distributed in the bounded domains. As the number of devices increases, we look for more refined allocation policy and the expected reward should be larger. 

\begin{figure}[ht]
  \centering
  \includegraphics[scale=.52]{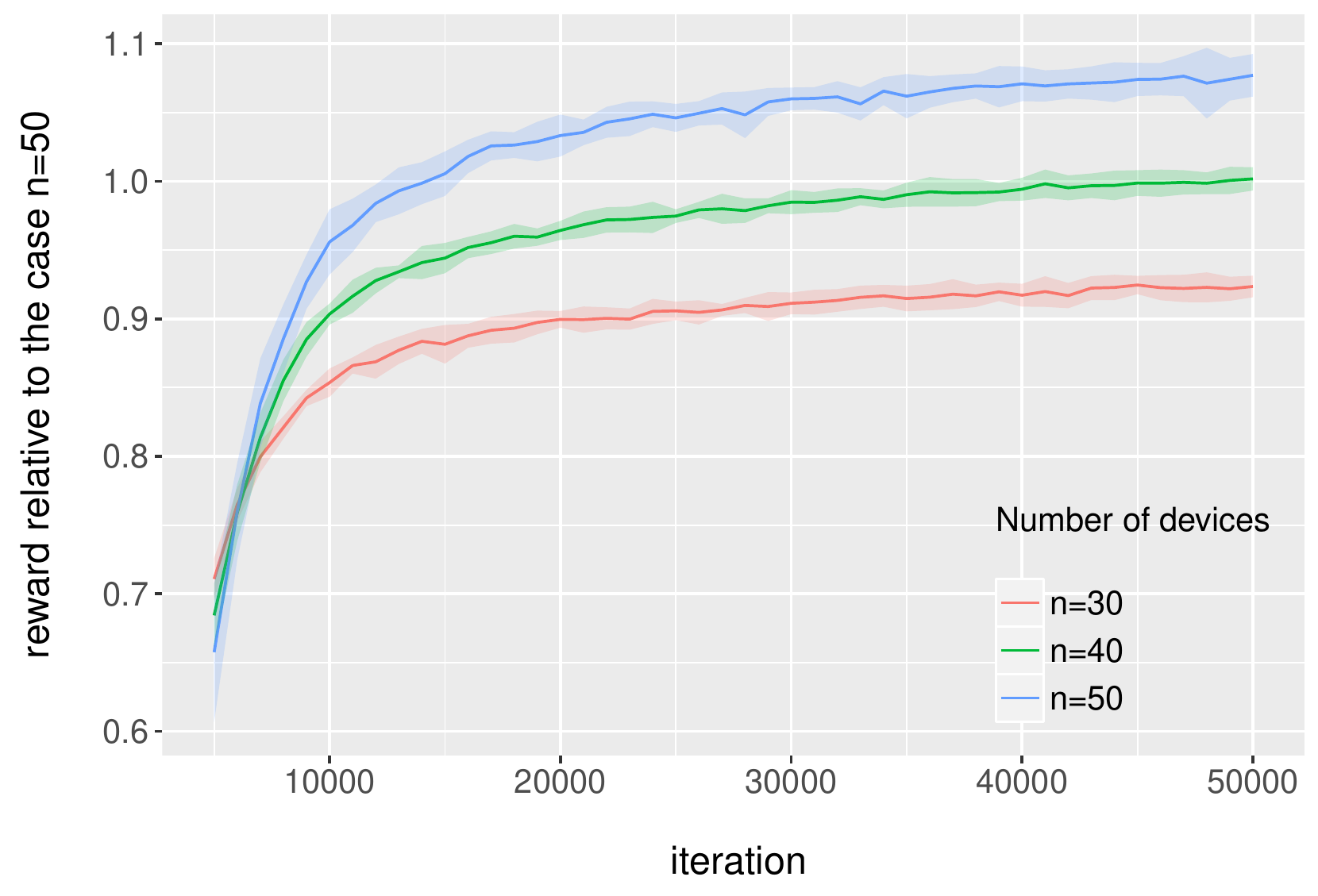}
  \caption{Reward relative to the expected total reward (with controls satisfying constraints strictly) in the case $n=50$ as a function of the number of iterations on validation samples. The shaded area depicts the mean $\pm$ the standard deviation over five different random seeds. The average relative reward on test samples is $0.926, 0.965$ for $n=30, 40$. The average running time for three cases is 6672~s, 8374~s and 10219~s.}
  \label{fig:multi_reward}
\end{figure}

We use the same learning parameters as in the case of single device except that we reduce all the penalty coefficients to $30$ and batch size to $64$. Learning curves plotted in Figure \ref{fig:multi_reward} confirms our expectation that the reward increases as the number of devises increases.
The learning curves behave similarly as in the case of a single device and different random initializations still give similar expected reward. Note that the function space of the control policy: $\mathbb{R}^{n+2}\rightarrow \mathbb{R}^{3n}$ grows as $n$ increases from 30 to 50, while our algorithm still finds near-optimal solution with slightly increased computational time.

\section{Conclusion}
In this paper, we formulate a deep learning approach to directly solve high-dimensional stochastic control problems in finite horizon. We use feedforward neural networks to approximate the time-dependent control at each time and stack them together through model dynamics. The objective function for the control problem plays the role of the loss function in deep learning.  Our numerical results suggest that for different problems, even in the presence of multiple constraints, this algorithm finds near-optimal solutions with great extendability to high-dimensional case.

The approach presented here should be applicable to a wide variety of
problems in different areas including dynamic resource allocation with many
resources and demands, dynamic game theory with many agents and wealth
management with large portfolios.
In the literature these problems were treated under different assumptions such 
as separability or mean-field approximation.
As suggested by the results of this paper, the
deep neural network approximation
should provide a more general setting and should give better results.


\begin{small}
\bibliographystyle{unsrt}
\bibliography{DPwithDL}
\end{small}

\end{document}